\documentclass{article}

\usepackage[preprint]{nips_2018}

\usepackage[utf8]{inputenc} 
\usepackage[T1]{fontenc}    
\usepackage{hyperref}       
\usepackage{url}            
\usepackage{booktabs}       
\usepackage{amsfonts}       
\usepackage{nicefrac}       
\usepackage{microtype}      
\usepackage{graphicx}
\usepackage{listings}

\title{Training Neural Speech Recognition Systems with Synthetic Speech Augmentation}

\author{
    Jason Li, Ravi Gadde\thanks{work done while the author was an intern at NVIDIA} , Boris Ginsburg, Vitaly Lavrukhin \\
  NVIDIA \\
  Santa Clara, CA 95051\\
  \texttt{\{jasoli,rgadde,bginsburg,vlavrukhin\}@nvidia.com} \\
}

\begin{document}

\maketitle

\begin{abstract}

  Building an accurate automatic speech recognition (ASR) system requires a large dataset that contains many hours of labeled speech samples produced by a diverse set of speakers. The lack of such open free datasets is one of the main issues preventing advancements in ASR research. To address this problem, we propose to augment a natural speech dataset with synthetic speech. We train very large end-to-end neural speech recognition models using the LibriSpeech dataset augmented with synthetic speech. These new models achieve state of the art Word Error Rate (WER) for character-level based models without an external language model.
   
\end{abstract}

\section{Introduction}
\label{intro}

There has been a large amount of success in using neural networks (NN) for automatic speech recognition (ASR). Classical ASR systems use complicated pipelines with many heavily engineered processing stages, including specialized input features, acoustic models, and Hidden Markov Models (HMMs). Deep NNs have traditionally been used for acoustic modeling \citep{Waibel1989, Hinton2012}. A key breakthrough occurred when a state of the art ASR system, Deep Speech, was built using an end-to-end deep learning approach \citep{DeepSpeech2014}. This model replaced both acoustic modeling, and HMMs with very deep neural networks and was able to directly translate spectrograms into English text transcriptions. This end-to-end approach was extended in follow-up papers \citep{Amodei2016DeepS2,collobert2016wav2letter}. Recent advances in ASR have further improved upon these models using even more advanced techniques such as replacing n-gram language models with a neural language model, usually in the form of a recurrent neural network (RNN) \citep{ASRAttention2018, Povey2018SemiOrthogonalLM, CAPIO2017}. 

As opposed to making neural networks more complex, we were interested in an orthogonal direction of study: whether we can improve quality by creating larger models. In order to train such large models, deep NNs require a vast quantity of data to be available. However the lack of a large public speech dataset blocked us from successfully building large NN models for ASR.
We were inspired by recent work in improving translation systems using synthetic data \citep{FacebookGERENGBackTranslation}, and as such, we decided to augment speech with synthetic data.
\footnote{
Synthesized speech has previously been used to improve speech recognition for low-resource languages \citep{Rygaard2015}.
}
Furthermore, given the recent impressive improvement in neural speech synthesis models \citep{WaveNet, shen2018natural}, it becomes cheap to generate high quality speech with varying prosody.

We show that by using synthetic speech created from a neural speech synthesis model, we can improve ASR performance compared to models trained using only LibriSpeech data \citep{panayotov2015librispeech}. By naively increasing the depth of the model, we observe that the synthetic data allows us to achieve state of the art WER using a greedy decoder.

\section{Synthetic Speech Dataset}
\label{creation}
We use the Tacotron-2 like model from the OpenSeq2Seq 
\footnote{We used OpenSeq2Seq  both to create a synthetic speech dataset and build ASR.
} toolkit \citep{OpenSeq2Seq} and add Global Style Tokens (GST) \citep{GlobalStyleTokens} to learn multiple speaker identities. Tacotron-2 with GST (T2-GST) was trained on the MAILABS English-US dataset \citep{mailabs}
with approximately 100 hours of audio recorded by 3 speakers. T2-GST was able to learn all 3 different speaking styles and different accents that they use to portray different characters.

Using the T2-GST model, we created a fully synthetic version of the LibriSpeech training audio. In order to produce audio, T2-GST requires a spectrogram used for the style and the audio transcription. For the transciption, we took the transcripts from the train-clean-100, train-clean-360, and train-other-500 LibriSpeech splits and randomly paired them with style spectrograms from the MAILABS English-US dataset. At the end, we had a dataset that was the same size as the original training portion of the LibriSpeech dataset but spoken in the tones of the speakers from the MAILABS  dataset.

The audio from the T2-GST model could be further controlled by the amount of dropout in the prenet of the decoder. By decreasing the dropout rate, we found that we could slightly distort the audio. The main difference that we noticed was that the lower the dropout rate, the faster the resulting audio would sound. Using this observation, we further increased the size of the synthetic dataset. Thus, we used 46\%, 48\%, and 50\% for the dropout and created a synthetic speech dataset that was 3 times as large as the LibriSpeech training dataset.

\section{Training Speech Recognition with Synthetic Data}
\label{results}

\subsection{Neural Speech Recognition Models}
Our speech recognition model is an end-to-end neural network that takes logarithmic mel-scale spectrograms as inputs and produces characters. We use a deep convolutional NN model, which we will further address as Wave2Letter+ (w2lp) 
\footnote{More model details:  
\href{OS2S}{https://nvidia.github.io/OpenSeq2Seq/html/speech-recognition/wave2letter.html}
}
. It is based on Wav2Letter \citep{collobert2016wav2letter} except:
\begin{itemize}
  \item We use ReLU instead of Gated Linear Unit
  \item We use batch normalization instead of weight normalization 
  \item We add residual connections between convolutional blocks
  \item We use Connectionist Temporal Classification loss instead of Auto Segmentation Criterion
  \item We use Layer-wise Adaptive Rate clipping (LARC) for gradient clipping
\end{itemize}

\begin{figure*}[!h]
\includegraphics[width=14cm]{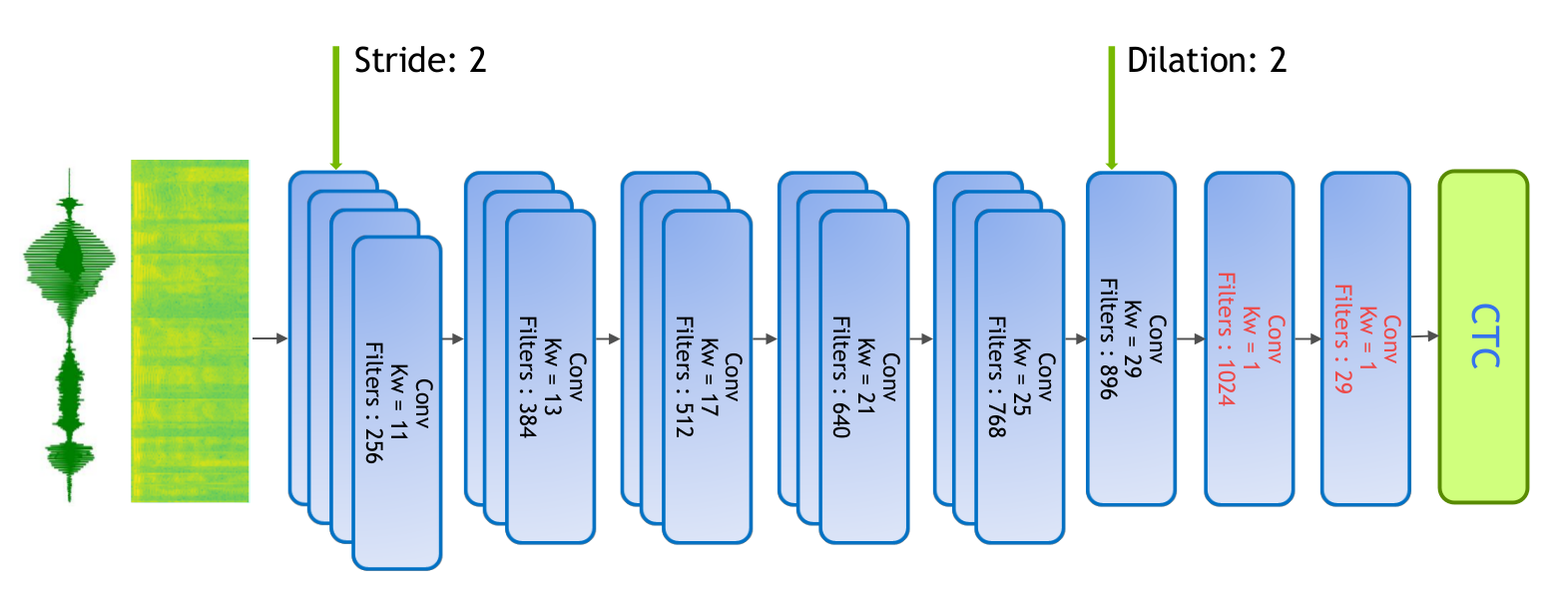}
\caption{\label{fig:w2lplus} Base Wave2Letter+ model.}
\end{figure*}

\newpage

The base Wave2Letter+ model has 19  convolutional layers:
\begin{itemize}
  \item 1 pre-processing layer at the beginning of the network
  \item 15 layers consists of 5 blocks of 3 repeating convolutional layers
  \item 3 post-processing layers at the end of the network 
\end{itemize}
We made the base model deeper and experimented with 24, 34, 44, and 54 layer networks. The 24 layer networks consists of 5 blocks of 4 repeating convolutional layers. The 34, 44, and 54 layer network consists of 10 blocks of 3, 4, and 5 repeating layers respectively.

\subsection{Word Error Rate Improvement using Synthetic Augmentation}
The training dataset was created by combining the synthetic data with the original training data from LibriSpeech. Despite the synthetic data being larger than the natural data, we found it most helpful to sample natural and synthetic data at a 50/50 rate.

Our best performing model, the 54 layer network, currently has a word error rate of 4.32\% on test-clean and a word error rate of 14.08\% on test-other by greedily choosing the most probable character at each step without any language model rescoring. To the best of our knowledge, the previous best performing model without using any language model achieves 4.87\% on test-clean and 15.39\% on test-other \citep{ASRAttention2018}. The complete results can be found in Table \ref{LibriVsSyn}. Models trained on the combined dataset outperform those trained on original LibriSpeech. The augmented dataset improves results on test-clean by 0.15, and 0.44 for the 24 and 34 layer models. For test-other, we see an improvement of 0.08, and 0.74 for the 24 and 34 layer models.

By using beam search with beam width 128 and the 4-gram OpenSLR \footnote{The LM can be found here: www.openslr.org/11} language model rescoring, we improved our WER on test-other to 12.21\% on the 54 layer model which is better than previous public 4-gram language models and comparable to LSTM language models \citep{ASRAttention2018}.

\begin{table}[!h]
\centering
\begin{tabular}{||c|c|c|c|c|c||} 
 \hline
 Model & Dataset Used & \multicolumn{2}{c|}{Dev} & \multicolumn{2}{c||}{Test} \\
 & & Clean & Other & Clean & Other \\
 \hline\hline
 attention-Zeyer et al. & LibriSpeech & 4.87 & 14.37 & 4.87 & 15.39 \\
 \hline
 w2lp-24 & LibriSpeech & 5.44 & 16.57 & 5.31 & 17.09 \\
 w2lp-24 & Combined & 5.12 & 16.25 & 5.16 & 17.01\\ 
  \hline
 w2lp-34 & LibriSpeech & 5.10 & 15.49 & 5.10 & 16.21 \\
 w2lp-34 & Combined & 4.60 & 14.98 & 4.66 & 15.47 \\
  \hline
 w2lp-44 & Combined & 4.24 & 13.87 & 4.36 & 14.37 \\
 w2lp-54 & Combined & 4.32 & 13.74 & 4.32 & 14.08 \\ [1ex] 
 \hline
\end{tabular}
\caption{Greedy WER on LibriSpeech for Different Models and Datasets}
\label{LibriVsSyn}
\end{table}

\subsection{How To Mix Natural and Synthetic Speech}
We performed a number of additional experiments to find the best sampling ratio between synthetic data and LibriSpeech. We tested training on only LibriSpeech, a 50/50 split, a 33/66 split, and a pure synthetic dataset. All tests were done on the 34 layer model. The results are shown in Table \ref{SampleProb}.

\begin{table}[!h]
\centering
\begin{tabular}{||c|c|c|c|c|c||}
 \hline
 Model & Sampling Ratio & \multicolumn{2}{c|}{Dev} & \multicolumn{2}{c||}{Test} \\
 & (Natural/Synthetic) & Clean & Other & Clean & Other \\
 \hline\hline
 w2lp-34 & Natural & 5.10 & 15.49 & 5.10 & 16.21 \\ 
 w2lp-34 & 50/50 & 4.60 & 14.98 & 4.66 & 15.47 \\
 w2lp-34 & 33/66 & 4.91 & 15.18 & 4.81 & 15.81 \\
 w2lp-34 & Synthetic & 51.39 & 80.27 & 49.80 & 81.78 \\ [1ex] 
 \hline
\end{tabular}
\caption{Greedy WER for Different Ratios Between Natural and Synthetic Datasets}
\label{SampleProb}
\end{table}

Despite the larger amount of synthetic data, the synthetic dataset fails to capture the larger variety of LibriSpeech. We believe that this effect could be moderated if a speech synthesis model with larger speaker variety was used as opposed to the current 3 speaker speech synthesis model. A 50/50 split between the natural and synthetic seems to be a good ratio for our dataset. 

\subsection{Traditional Speech Augmentation vs Synthetic Speech}
Adding synthetic data proved to be better regularization than standard regularization techniques. In addition to dropout which is employed for all models, OpenSeq2Seq supports speech augmentation such as adding noise and time stretching. Using these 3 techniques, we tested 4 additional models. We tested 2 larger dropout factors, and on top on this, we tested whether speech augmentation would improve performance. Since the dropout factor varies by layer, we multiply the local dropout keep probabilities by a global dropout keep factor. All tests were done on the 34 layer model. The tests and results are detailed in Table \ref{reg}.

A slightly larger dropout resulted in minor improvement in WER. The effects of speech augmentation seem to be negligible or, in the case of large dropout, make WER worse. Adding synthetic data significantly outperforms all other methods of regularization.

\begin{table}[!h]
\centering
\begin{tabular}{||c|c|c|c|c|c|c|c||} 
 \hline
 Model &  Dropout  & Time Stretch & Noise & \multicolumn{2}{c|}{Dev} & \multicolumn{2}{c||}{Test} \\
      & Keep Factor & Factor & (dB) & Clean & Other & Clean & Other \\
 \hline\hline
 LibriSpeech & None & None & None & 5.10 & 15.49 & 5.10 & 16.21 \\
 \hline

 Dropout     & 0.9 & None & None & 5.01 & 15.15 & 5.15 & 15.70 \\
 Dropout + Aug & 0.9 & 0.05 & [-90, -60] & 5.07 & 15.00 & 5.02 & 15.83 \\
 \hline
 Dropout     & 0.75 & None & None & 5.46 & 15.77 & 5.39 & 16.62 \\
 Dropout + Aug &  0.75 & 0.1 & [-90, -60] & 5.80 & 16.33 & 5.72 & 17.41 \\
 \hline
 Combined & None & None & None & 4.60 & 14.98 & 4.66 & 15.47 \\
 \hline
\end{tabular}
\caption{Greedy WER on Using Different Regularization Techniques}
\label{reg}
\end{table}

\section{Conclusion and Future Plans}
Using synthetic data is an effective way to build large neural speech recognition systems. The synthetic data should be combined with the natural data in the correct ratio to obtain best results. With this method, we achieved a WER of 4.32\% on test-clean and a WER of 14.08\% on test-other using a greedy decoder. This is the current state of the art on character-level based greedy decoding. Furthermore, using a language model and a beam search width of 128, we get 12.21\% WER on test-other. 

For future studies, we are interested in creating a larger synthetic dataset with noise. For now, we have restricted ourselves to take transcriptions from the training subsets of LibriSpeech, but the speech synthesis models are general enough to accept any transcript. It would be interesting to scrape text from another source and add to the training set additional phrases not found in LibriSpeech.

\subsubsection*{Acknowledgments}
We would like to thank Rafael Valle, Ryan Leary, Igor Gitman, Oleksii Kuchaiev and Ujval Kapasi for helpful conversation and advice throught this project. 

\bibliography{nips_2018}
\bibliographystyle{acl_natbib}
\end{document}